\documentclass[10pt,twocolumn,letterpaper]{article}

\usepackage[pagenumbers]{cvpr} 

\usepackage{amsfonts}
\usepackage{algorithm}
\usepackage{algorithmic}
\usepackage{multirow}
\usepackage{placeins} 

\definecolor{cvprblue}{rgb}{0.21,0.49,0.74}
\usepackage[pagebackref,breaklinks,colorlinks,allcolors=cvprblue]{hyperref}

\def\paperID{29} 
\def\confName{CVPR}
\def\confYear{2026}

\title{ Weight Group-wise Post-Training Quantization for Medical Foundation Model}

\newcommand{\authorcell}[2]{%
  \parbox[t][#1][t]{0.31\linewidth}{\centering #2}%
}

\author{
\centering
\begin{tabular}{@{}ccc@{}}
\authorcell{6em}{Yineng Chen \thanks{Equal Contribution}\\
University at Albany, SUNY\\
{\tt\small ychen77@albany.edu}}
&
\authorcell{6em}{Peng Huang\footnotemark[1] \thanks{Visiting researcher at University at Albany, SUNY.} \\
University at Albany, SUNY\\
Southwest Jiaotong University\\
{\tt\small phuang@albany.edu\\ huangpeng@my.swjtu.edu.cn }}
&
\authorcell{4.5em}{Aozhong Zhang\\
University at Albany, SUNY\\
{\tt\small azhang3@albany.edu}}
\\
\authorcell{6em}{Hui Guo\\
University at Albany, SUNY\\
{\tt\small hguo@albany.edu}}
&
\authorcell{4.5em}{Penghang Yin\\
University at Albany, SUNY\\
{\tt\small pyin@albany.edu}}
&
\authorcell{4.5em}{Shu Hu\\
Purdue University\\
{\tt\small hu968@purdue.edu}}
\\ [-1em]
\authorcell{4.5em}{Shao Lin\\
University at Albany, SUNY
{\tt\small slin@albany.edu}}
&
\authorcell{4.5em}{Xin Li\\
University at Albany, SUNY
{\tt\small xli48@albany.edu}}
&
\authorcell{4.5em}{Tzu-Jen Kao\\
GE HealthCare\\
{\tt\small kao@gehealthcare.com}}
\\
\authorcell{5em}{Balakrishnan Prabhakaran\\
University at Albany, SUNY\\
{\tt\small bprabhakaran@albany.edu}}
&
\authorcell{5em}{MingChing Chang\\
University at Albany, SUNY\\
{\tt\small mchang2@albany.edu}}
&
\authorcell{5em}{Xin Wang\thanks{Corresponding author.}\\
University at Albany, SUNY\\
{\tt\small xwang56@albany.edu}}
\end{tabular}
}

\begin{document}
\maketitle
\begin{abstract}
Foundation models have achieved remarkable results in medical image analysis. 
However, its large network architecture and high computational complexity significantly impact inference speed, limiting its application on terminal medical devices. 
Quantization, a technique that compresses models into low-bit versions, is a solution to this challenge. 
In this paper, we propose a post-training quantization algorithm, Permutation-COMQ. 
It eliminates the need for backpropagation by using simple dot products and rounding operations, thereby removing hyperparameter tuning and simplifying the process. Additionally, we introduce a weight-aware strategy that reorders the weight within each layer to address the accuracy degradation induced by channel-wise scaling during quantization, while preserving channel structure. Experiments demonstrate that our method achieves the best results in 2-bit, 4-bit, and 8-bit quantization. 
\end{abstract}    
\section{Introduction}
\label{sec:intro}

Foundation models, large artificial intelligence (AI) models pre-trained on large, diverse, and typically unlabeled datasets, have shown significant promise in image analysis. 
A pioneering foundation model based on prompt-based segmentation is the \textbf{S}egment \textbf{A}nything \textbf{M}odel (SAM), which has achieved significant breakthroughs in image segmentation after being trained on approximately 11 million images \cite{SAM}. However, this model exhibits substantial limitations when dealing with objects with weak boundaries or low-contrast targets \cite{huang2025diffusion}. Medical images differ fundamentally from natural images in both acquisition mechanisms and statistical properties. They often exhibit low contrast, high inter-region similarity, and subtle intensity variations across organs, lesions, and tissues \cite{asgari2021deep,duncan2002medical}. To address these challenges, MedSAM was developed by training a medical-specific version of SAM using more than 1.5 million medical image masks \cite{ma2024medsam}. Despite the emergence of highly advanced foundation models in medical imaging, medical images inherently possess higher resolution and richer information than natural images. This distinction is particularly pronounced in high-dimensional modalities, such as CT and MRI scans, which significantly increase computational demands. Moreover, low-latency and high-precision inference is critical in real-time applications such as surgical navigation and lesion detection. The billions or even trillions of parameters that models such as MedSAM have severely impacted their inference speed and application to terminal medical devices. Reducing storage requirements, minimizing memory usage, and lowering computational costs have become significant challenges in the clinical application of medical foundation models.

Several techniques have been explored to alleviate these challenges, including network pruning, knowledge distillation, and numerical quantization \cite{hohman2024model,menghani2023efficient}. Among these approaches, quantization is particularly attractive because it reduces the computational and memory requirements of foundation models by representing model parameters with lower-precision data types, while preserving the original model architecture. By mapping floating-point values to lower-precision integer representations, such as int8, quantization significantly reduces memory usage and can accelerate inference. To address the pressing need for the effective deployment of large models in the healthcare domain, we have designed a coordinate-wise minimization quantization method (Permutation-COMQ). This algorithm significantly reduces both computational and storage requirements in the absence of adequate computational and training resources, providing strong support for the application of foundation models in clinical practice. 

The major contributions are as follows:

\begin{enumerate}
    \item We propose a \textbf{P}ost-\textbf{T}raining \textbf{Q}uantization (PTQ) algorithm Permutation-COMQ. 
    This algorithm eliminates the need for model fine-tuning and achieves quantization only through dot products and rounding operations, offering a cost-effective quantization solution for large medical models.
    
    \item Permutation-COMQ optimizes the weights by minimizing a series of univariate quadratic functions, avoiding the complexity associated with backpropagation and the computation of the Hessian matrix inverse, thereby enhancing quantization efficiency and simplifying the optimization process. 
    
    \item Permutation-COMQ introduces a post-permutation scaling scheme to address the varying distributions of different-sized weights in the original weight matrix, thereby mitigating the PTQ accuracy loss caused by channel-wise scaling during the quantization process.
    
\end{enumerate}

\begin{figure*}[t]
\centerline{\includegraphics[width=0.85\linewidth]{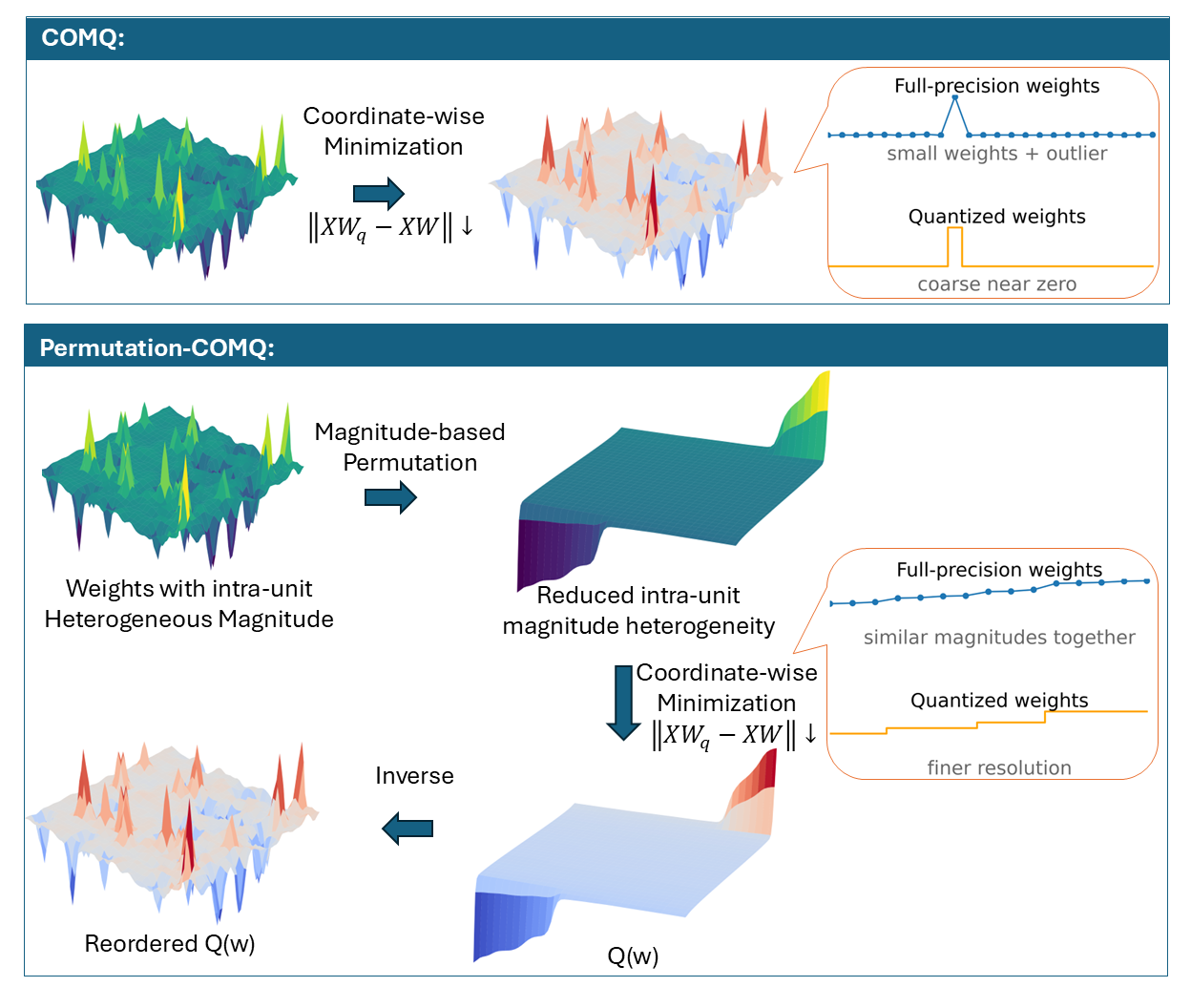}}
\caption{Conceptual illustration of COMQ and Permutation-COMQ
\label{fig:workflow}
\textbf{Top:} COMQ has quantization scale factors dominated by outliers due to heterogeneous weight magnitudes within each quantization unit, leading to coarse quantization of small weights. 
\textbf{Bottom:} Permutation-COMQ reorders weights by magnitude to group similar values, reducing intra-unit magnitude heterogeneity resulting in finer quantization resolution. The quantized weights are then mapped back to the original order via inverse permutation. \\}
\label{fig:summary}
\end{figure*}

\section{Related Works}
\label{sec:RelatedWorks}

\subsection{Medical Foundation Model}
Foundation models are large models pre-trained on large and diverse data, with the goal of supporting a wide range of downstream tasks. Recent works have explored the use of foundation models in various medical imaging tasks such as segmentation, classification and registration \cite{tsai2024uu, huang2024robustly, hu2025improving, zheng2024contextual, tsai2024uu2, lin2024robust,zhu2024cgd, hu2024umednerf, wang2024neural}. Among these tasks, segmentation has received particular attention due to its fundamental role in clinical analysis and decision-making.

SAM has demonstrated powerful cross-task segmentation capabilities, enabling a unified paradigm for medical imaging across modalities and lesion types. To improve the adaptability of SAM to medical scenarios, MedSAM, Med2D, and Medical SAM Adapter etc. fine-tune these models on additional medical images to improve segmentation precision on medical images \cite{ma2024medsam,cheng2023sam2d,chen2024sam2}. Several extensions of MedSAM have been proposed, such as U-MedSAM, and AutoMedSAM \cite{wang2024u, huang2025diffusion}. Meanwhile, the complex model structure hinders its utilization in clinical settings. In response, models like FastSAM and MobileSAM have simplified the model architecture and prompt schemes \cite{zhao2023fast,zhang2023mobilesamv2}. However, there remains a significant gap between these models and their practical deployment. Therefore, it is important to compress these models for the deployment of smart healthcare on medical terminal devices.

\subsection{Quantization}
Deep learning models have relatively centralized resources during training. However, in applications, terminal devices cannot meet inference requirements under resource-constrained conditions. Quantization reduces the model's storage and computational demands by mapping high-precision floating-point weights and activations to low-bit representations \cite{zhang2024comq,zhang2024magr}. 
Quantization methods can be broadly categorized into two main types: \textbf{Q}uantization-\textbf{A}ware \textbf{T}raining (QAT) and PTQ. 
QAT incorporates quantization effects directly into the training process by simulating the effects of low-precision arithmetic during training. Specifically, fake quantization operations are inserted into the computation graph to emulate the behavior of quantized weights and activations (e.g., at 8-bit or 4-bit precision). This allows the model to learn parameters that are robust to reduced numerical precision, resulting in improved performance during quantized inference \cite{nagel2021white}. However, QAT typically requires access to large-scale training data and substantial computational resources, which are often unavailable or impractical in medical imaging scenarios. 
In contrast, post-training quantization (PTQ) does not require retraining the model and can be applied in a data-free manner or with a small unlabeled calibration dataset. PTQ directly converts pretrained full-precision weights (le.g., FP32) into lower-bit representations (e.g., INT8), making it particularly appealing for deploying large medical foundation models under data privacy and computational constraints \cite{nagel2021white,lang2024comprehensive}. 
Various PTQ techniques have been developed. 
RTN (Round-to-Nearest) is a straightforward post-training quantization technique that scales full-precision weights and rounds the scaled values to the nearest integer \cite{kogan2025selective}. Due to its simplicity, RTN has been widely adopted as a standard quantization baseline \cite{lin2024awq,frantar2022gptq,tang2023easyquant}. 
GPTQ uses approximate second-order information to minimize the output error, while compensating for that error by updating remaining weights. It has demonstrated strong performance in quantizing models down to 3–4 bit precision with minimal accuracy degradation \cite{frantar2022gptq}. 
COMQ formulates quantization as a sequence of optimization problems aiming to minimize the output error caused by quantizing each weight. However, COMQ can be sensitive to weight distributions with large dynamic ranges, where extreme values dominate the scaling factors and limit quantization resolution for the majority of weights \cite{zhang2024comq}.

\noindent {\bf Granularity in quantization} describes how quantization parameters, such as scale and zero-point, are shared across a tensor \cite{krishnamoorthi2018quantizing,nagel2021white}. \textbf{Per-layer quantization} applies a single set of quantization parameters to all values within a layer, offering simplicity and high computational efficiency. However, per-tensor quantization is sensitive to outliers and may suffer from precision degradation when different channels exhibit heterogeneous value distributions. In contrast, \textbf{per-channel quantization} assigns independent quantization parameters for each output channel, enabling better accommodation of channel-wise variations and generally achieving higher accuracy in PTQ settings, particularly for deep and over-parameterized models commonly used in medical image analysis, albeit at the cost of increased computational overhead and implementation complexity. In addition, \textbf{block-wise quantization} divides channels into small blocks, such as 64 contiguous elements within a channel, and a single set of quantization parameters is shared within each group. However, the choice of block size is a trade-off between simplicity and accuracy, with a smaller block size means more scale factors and increased memory overhead.

\noindent {\bf Uniform PTQ} 
maps input values evenly to the range of quantized values.
A set of points $ w\in \mathbb{R}^m$ is quantized using a quantization step $\delta = \frac{\max(w) -\min(w)}{2^b-1}$ and zero-point $z = \left\lfloor\frac{\min(w)}{\delta} \right\rceil$. The process maps $w$ onto a discrete set of scaled integer grid points: $\mathbf{Q}=\{z \cdot \delta, (z+1)\cdot\delta,\ldots,\left(z+(2^{b}-1)\right)\cdot \delta \}^m$. The  quantized values $w_q$ are obtained through the following transformation:
\[
    w_q = \delta \cdot \left (\text{clamp} \left(\left\lfloor\frac{{w}}{\delta} \right\rceil -z, 0, 2^{b}-1 \right) + z\right).
\]
For per-channel post-training quantization (PTQ), the quantization step $\delta$ is typically determined based on the minimum and maximum values of the pre-trained weight tensor $W$ within each channel. This fixed step size ensures consistent numerical representation throughout quantization.
\section{Proposed Method}
\label{sec:Method}

\subsection{Preliminaries}
\noindent {\bf Notations}
We represent vectors using bold lowercase letters and matrices using bold uppercase letters. For any matrix $ \boldsymbol{X} \in \mathbb{R}^{m\times n}$, its transpose is denoted as $\boldsymbol{X}^\top \in \mathbb{R}^{n\times m}$. The Frobenius norm of $\boldsymbol{X}$ is defined as $\|\boldsymbol{X}\|_{\mathrm{F}}= \sqrt{\sum_{i=1}^m\sum_{j=1}^n X_{i,j}^2}$. 
Additionally, for vectors $\boldsymbol{x}$ and $\boldsymbol{y}$, the Hadamard (element-wise) product is defined as $\boldsymbol{x} \odot \boldsymbol{y} := (x_1 y_1, \dots, x_n y_n)\in\mathbb{R}^n$. This definition extends similarly to the element-wise product of two matrices.

\subsection{Overview}

Permutation-COMQ is a weight-aware coordinate descent quantization framework that reorders weight matrix before solving for quantized weights. Our objective is to find quantized weights $W_q$ that minimize the following function.

\begin{equation} \label{eq:minWi}
    \min_{\boldsymbol{W_q}   \in \mathcal{W}} \; \| \boldsymbol{x} \boldsymbol{W_q} - \boldsymbol{x} \boldsymbol{W} \|^2,
\end{equation}

where $\boldsymbol{x}$ denotes input matrix and $\boldsymbol{W}$ represents the full-precision weights. 

Following the COMQ formulation, we solve this optimization problem using a coordinate descent algorithm, treating each quantized weight and its associated scaling factor as optimization coordinates. \cite{zhang2024comq} The multivariate optimization problem is decomposed into a sequence of univariate subproblems, where one coordinate is updated at a time while keeping all others fixed.

Another challenge arises when weight values within a channel exhibit large dynamic ranges or heterogeneous magnitude distributions. In such a case, the estimated scaling factors may be dominated by outliers, leading to an increased quantization error. To address this issue, we introduce a permutation step prior to optimization. Specifically, the weight matrix $\boldsymbol{W} \in \mathbb{R}^{m\times n}$ is permuted based on magnitude ordering to obtain a permuted matrix $\boldsymbol{W_p} \in \mathbb{R}^{m\times n}$. This permutation redistributes weights such that elements within each quantization unit exhibit more homogeneous magnitude distributions. Coordinate-wise minimization is then applied to the permuted matrix. After quantization, the inverse permutation restores the original weight ordering. Figure \ref{fig:workflow} demonstrates the workflow of the proposed permutation-COMQ.

\subsection {\bf Weight-aware Optimization for Permutation-COMQ}

\textbf{Permutation}
To enhance quantization performance, we introduce a weight-aware sorting operation applied within each layer that strategically restructures the weight matrix for improved numerical properties. Given a weight matrix $\boldsymbol{W}$, we define a permutation matrix $\boldsymbol{P}$ that reorders the elements of $\boldsymbol{W}$ in ascending order. The transformation is applied as follows:
\[\boldsymbol{W_p} = \boldsymbol{P} \boldsymbol{W} \]
The resulting matrix $\boldsymbol{W_p} $ is structured such that values of similar magnitude are clustered into localized regions, enhancing its compatibility with quantization. We then perform coordinate-wise quantization directly on $\boldsymbol{W_p} $, yielding the quantized matrix $\boldsymbol{W_q}$.

\textbf{Reverse}
The permutation is applied only to facilitate quantization. To restore the original structure, we apply the inverse permutation after quantization, reconstructing the quantized weights as:
\[\widetilde{\boldsymbol{W}}_q = \boldsymbol{P}^{-1}\boldsymbol{W}_q\]
This transformation significantly refines the weight distribution, making it more conducive to affine uniform quantization by reducing local quantization error and preserving intra-group coherence. By minimizing variance within quantization clusters and maintaining the relative consistency of weight values, this method effectively suppresses quantization-induced distortions and enhances the precision of low-bit representations. Consequently, it enables a more efficient quantization process, ultimately leading to improved model accuracy and overall performance.

\noindent {\bf Per-channel Coordinate-wise Minimization} 
Specifically, the quantized weight matrix $\boldsymbol{W_q}\in \mathbb{R}^{m\times n}$ is expressed as
\[
     \boldsymbol{W_q} = \boldsymbol{Q} \odot \boldsymbol{\delta}^\top= (\delta_1 \boldsymbol{q}_1,\ldots,\delta_n \boldsymbol{q}_n).
\]
where $ \boldsymbol{Q} \in\mathbb{S}^{m\times n}$ is the integer bit-code matrix for $\boldsymbol{W}_q$, $\boldsymbol{q_j}$ denotes the $j$-th column of $\boldsymbol{Q}$, and $\delta_j$ is the scale factor  associated with the the $j$-th column of the weight matrix $\boldsymbol{W}$. Consequently, the optimization problem \eqref{eq:minWi} can be reformulated as

 \begin{equation} \label{eq:minqi channel}
     \delta_j,\boldsymbol{q_j} = \arg \min_{\delta_j,\boldsymbol{q}_j} \|\delta_j \boldsymbol{Xq}_j -  \boldsymbol{Xw_j} \|^2.
 \end{equation}
 
 Let $\boldsymbol{\delta}^{k-1}\in\mathbb{R}^n$ and $\boldsymbol{q}^{k-1}\in\mathbb{S}^{m\times n}$ be the scaling factors and bit-code matrix produced by $(k-1)$-th iteration. Or equivalently, suppose $\boldsymbol{w}_q^{k-1} = \boldsymbol{q}^{k-1} \odot \boldsymbol{\delta}^{k-1}$ is the current quantized weight matrix. The updated bit-code $\boldsymbol{Q}_{i,j}^{k}$ is given by 
\begin{equation} \label{eq:Q updateq channel}
   \boldsymbol{Q}^{k}_{i,j} = \text{clip} \left ( \left \lfloor \frac{\left< \boldsymbol{x}_{i}, \boldsymbol{s}^k_{i,j} \right>}{\delta_j^{k-1} \|\boldsymbol{x}_{i}\|^{2}} \right \rceil, z_j, z_j+2^{b}-1 \right ),
\end{equation}
 
where $\boldsymbol{s}_{i,j}^k = \boldsymbol{X} \boldsymbol{w_j} - \delta_j^{k-1}(\sum_{t=1}^{i-1} Q_{t,j}^k \boldsymbol{x_t} + \sum_{t=i+1}^{m} Q_{t,j}^{k-1} \boldsymbol{x_t})$, and $z_j$ is the zero-point for quantizing the $j$-th column.  $\boldsymbol{Q}_{i,j}^{k}$ is updated row-wise iteratively for {$i=1,\ldots,m$}. 

\[
\boldsymbol{U}^{k}_{0} = \boldsymbol{X}(\boldsymbol{W}-\boldsymbol{W}_q^{k-1}) 
\]

\[
\boldsymbol{U}^k_i = \boldsymbol{U}^k_{i-1} - {x}_{:,i}  \otimes (\boldsymbol{w}_{i,:} -\delta^{k-1} \odot q^{k-1}_{i,:} )
\]


\[
\tilde{q}^k_{i,:} =
\frac{
\left( U_k^i + x_{:,i} \otimes w_{i,:} \right)^\top x_{:,i}
}{
\delta^{k-1} x_{:,i}^\top x_{:,i}
}
\]

\begin{equation} \label{eq:q updateq channel}  
q^k_{i,:} = \mathrm{clip} \left(
\left[ \tilde{q}^k_{i,:} \right]  ,
 z, z + 2^b-1
\right)
\end{equation}

\[
\boldsymbol{U}^k_i = \boldsymbol{U}^k_{i} + \boldsymbol{x}_{:,i} \otimes (\boldsymbol{w}_{i,:} - \delta^{k-1} \odot q^{k}_{i,:} )
\]

where $\boldsymbol{w}_{i,:} \in \mathbb{R}^{n}$ and $\boldsymbol{q}_{i,:} \in \mathbb{R}^{n}$ the $i$-th row of $\boldsymbol{W}$ and $\boldsymbol{Q}$, respectively. 

The scaling factors are then updated:
 
\begin{equation} \label{eq:delta updateq channel}   
   \delta^{k}_{j} =  \frac{\left< \boldsymbol{Xq}^k_j, \boldsymbol{Xw}_j \right>}{\| \boldsymbol{Xq}^k_j \|^2 }
\end{equation}

\noindent

\subsection{Algorithm}

Per-channel quantization assigns each column of a weight matrix with its own scale factor, which yields smaller quantization errors. The workflow of applying Permutation-COMQ to to a single linear layer under per-channel quantization is summarized in Alg. \ref{alg:COMQ channel}. 

Given a pre-trained weight matrix $\boldsymbol{W}\in\mathbb{R}^{m \times n}$, a feature matrix $\boldsymbol{X}$, and the number of iterations $\boldsymbol{K}$. The algorithm start with the initialization of the scaling factors $\delta^0_j$ and quantized weights $\boldsymbol{Q}^0$. The scale factor is initialized as $\delta^0_j = \lambda \frac{max(\boldsymbol{w}_j)-min(\boldsymbol{w}_j)}{2^b-1}$, for some $0 \leq \lambda \leq 1$ to preventing quantizing majority of values to zero. The $\boldsymbol{q}^0_j$ is initialized as $\boldsymbol{w}_j$  

The algorithm performs $K$ iterations. In each iteration, it loops over each row {$i=1,\ldots,m$}. For each row, it updates the quantized weights coordinate-wise by updating each element of $\boldsymbol{Q}^k_{i,j}$ as in Equation \eqref{eq:q updateq channel}. After updating all rows, it updates the scaling factors $\delta^k$ based on Equation \eqref{eq:delta updateq channel}   .

After $\boldsymbol{K}$ iterations, the quantized weight matrix $\boldsymbol{W}_q$ is obtained. Then inverse permutation is applied to restore the original structure.

\begin{algorithm}[t]
    \caption{algorithm for the per-channel quantization of one linear layer}
    \label{alg:COMQ channel}
    \begin{algorithmic}[1] 
        \REQUIRE Pre-trained weights $\boldsymbol{W} \in \mathbb{R}^{m\times n}$, feature matirx $\boldsymbol{X}$, and iteration number $K$.
            \STATE \textbf{Initialize} $\boldsymbol{Q}^0 = \boldsymbol{W}$
            \STATE $\boldsymbol{W}_p = \boldsymbol{P} \boldsymbol{Q}^0$
            \FOR{$k=1,\ldots,K$ }
                \FOR{$i=1,\ldots,m$}
                     \STATE{Update the coordinates $\{q_{i,j}^k\}$ as in \eqref{eq:q updateq channel}}
                \ENDFOR
                    \STATE{Update the scaling factors $\delta^k$ as in \eqref{eq:delta updateq channel}}
            \ENDFOR
            \STATE Compute $\boldsymbol{W}_q = (\delta_1^K \boldsymbol{q}_1^K,\ldots,\delta_n^K \boldsymbol{q}_n^K)$
            \STATE $ \widetilde{\boldsymbol{W}}_q  = \boldsymbol{P}^{-1}\boldsymbol{W_q}$
        \ENSURE Quantized weight $\widetilde{\boldsymbol{W}}_q $.
    \end{algorithmic} 
\end{algorithm}

\section{Experiments}
\label{sec:Experiments}

\subsection{Experiments on Simulation Data}

\begin{figure}[t]
\centerline{\includegraphics[width=\linewidth]{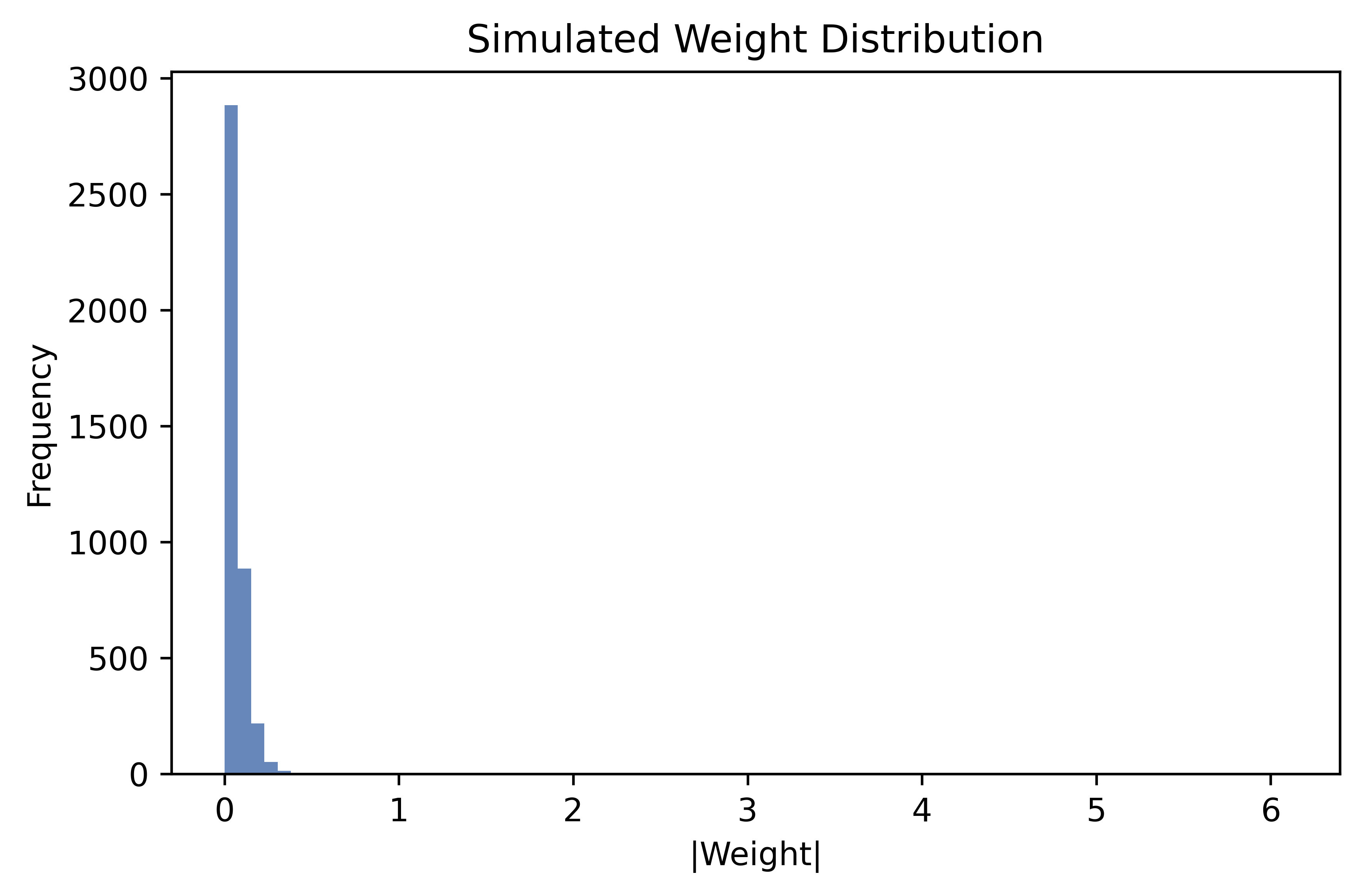}}
\caption{Distribution of magnitude of Simulated Weight.\\}
\label{fig:sim_w_hist}
\end{figure}

To illustrate the effect of permutation-COMQ on weight matrices with outliers, we apply COMQ and permutation-COMQ to simulated data under per-channel quantization at 8-, 4-, and 2-bit precision. We generated a synthetic weight matrix $W \in \mathbb{R}^{64\times 64}$ with sparse outliers to mimic realistic weight distributions. Specifically, base weights were sampled from a zero-mean Gaussian distribution with small variance. Structured variation was introduced by applying multiplicative row-wise and column-wise scaling factors. Finally, a small subset of entries was perturbed with large-magnitude values to simulate sparse outliers. As shown in Fig.~\ref{fig:sim_w_hist}, the distribution of the magnitude of the simulated weights was highly right-skewed, with the majority concentrated near zero while a few outliers reached magnitudes as large as 6. Such distributions were challenging for per-channel quantization, as the scale factors were determined by extreme values, resulting in poor resolution for the majority of weights. Calibration data $X \in \mathbb{R}^{256\times 64}$ were simulated from a standard Gaussian distribution with mild correlation induced by a linear mixing transformation. 

\begin{figure}[t]
\centerline{\includegraphics[width=\linewidth]{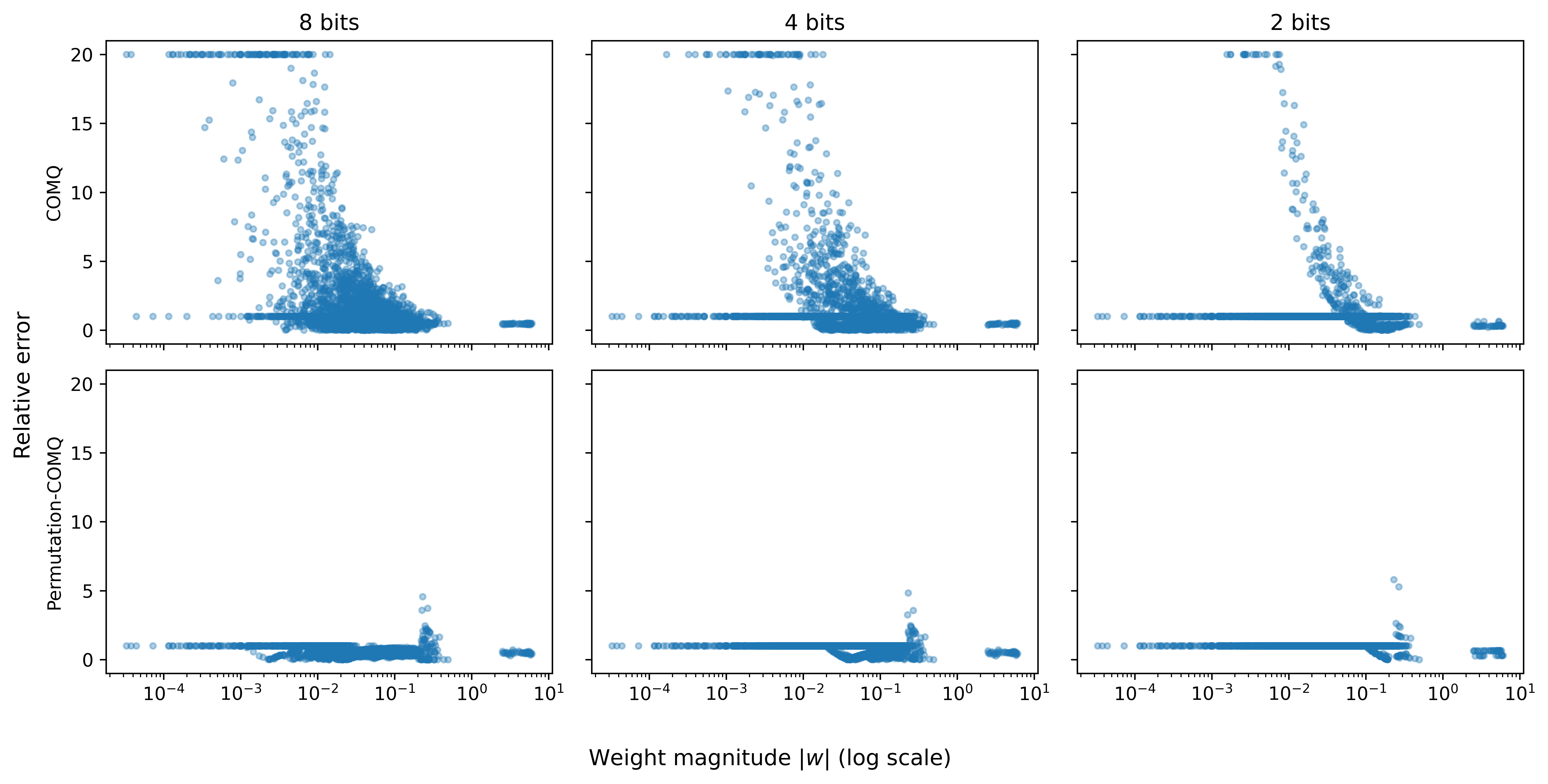}}
\caption{Relative quantization error for COMQ and permutation-COMQ under different bit-widths\\}
\label{fig:sim_rel_err}
\end{figure}

Fig.~\ref{fig:sim_rel_err} presents the relative error for COMQ and permutation-based COMQ under different bit-widths. Under COMQ, small-magnitude weights suffered disproportionately large relative error, while large weights exhibited relatively small relative error. This indicates that the scale factors were dominated by outliers, leading to coarse resolution to majority of weights. In contrast, permutation-COMQ reduced the relative error across a wider range of magnitudes, particularly for small weights, suggesting improved allocation of quantization scale factors and finer resolution to the majority of the weights.

\subsection{Experiments on Real Datasets}
\noindent{\textbf{Dataset}} We validate the performance of permutation-COMQ on AbdomenCT1K as the test dataset, which contains more than 1000 CT scans from 12 medical centers \cite{ma2021abdomenct}. The targets to be recognized include liver, kidney, spleen, and pancreas. These organs are of different shapes and sizes and can be effectively evaluated for the segmentation performance of the model.

\noindent{\textbf{Evaluation Metrics}} We employed the Dice Similarity Coefficient (DSC) and Normalized Surface Distance (NSD) to assess the segmentation performance. DSC is particularly effective in evaluating the overlap between the predicted mask and the ground truth. NSD focuses on aligning the boundaries of both. All experiments were conducted using PyTorch and tested on an NVIDIA A100-SXM4-80GB GPU.

\noindent{\textbf{Models}} We conduct experiments using MedSAM, a medical adaptation of the Segment Anything Model (SAM), which employs a Vision Transformer (ViT-B) image encoder along with a prompt-based mask decoder. In our experiments, ground-truth-derived bounding boxes are provided as prompts to guide the segmentation process, ensuring a fair evaluation of model performance under consistent conditions.

\begin{table}[t]
  \caption{Comparative Results on the AbdomenCT-1K Dataset. \textbf{Bold} indicates the best}
  \label{tab:ct}
  \centering
  \begin{tabular}{@{}cccc}
    \toprule
    Wbit & Method & DSC & NSD \\
    \midrule
    \multirow{3}{*}{2}  & COMQ \cite{zhang2024comq}  & 71.8 & 53.733\\
                        & RTN \cite{kogan2025selective}  & 29.79  & 30.221\\
                        & Ours  &  \textbf{86.939} & \textbf{78.935}  \\ \hline
    \multirow{3}{*}{4}  & COMQ \cite{zhang2024comq} & 91.874 & 90.011 \\
                        & RTN \cite{kogan2025selective} & 90.526 & 86.755\\
                        & Ours  & \textbf{93.434} & \textbf{93.089}  \\  \hline   
    \multirow{3}{*}{8}  & COMQ \cite{zhang2024comq} & 93.486 & 92.938  \\
                        & RTN \cite{kogan2025selective} & 93.499 & 92.936 \\
                        & Ours  & \textbf{93.615} & \textbf{93.204}  \\  \hline     
    32(BaseLine)        & -     & 93.505 & 92.969 \\
    \bottomrule
  \end{tabular}
\end{table}

\begin{table}[t]
  \caption{Results of ablation experiments for Per-Channel Weight-Aware}
  \label{tab:ab}
  \centering
  \begin{tabular}{@{}cccc}
    \toprule
    Wbit & Method & DSC & NSD \\
    \midrule
    \multirow{2}{*}{2}  & Per-Layer   & 74.46 & 54.825 \\
                        & Ours          & \textbf{86.939} & \textbf{78.935} \\ \hline
    \multirow{2}{*}{4}  & Per-Layer   & 55.439 & 43.685 \\
                        & Ours          & \textbf{93.434} & \textbf{93.089}  \\ \hline   
    \multirow{2}{*}{8}  & Per-Layer   & 93.282 & 92.461 \\
                        & Ours          & \textbf{93.615} & \textbf{93.204}  \\   
    \bottomrule
  \end{tabular}
\end{table}

\begin{figure*}[t]
\centerline{\includegraphics[width=0.98\linewidth]{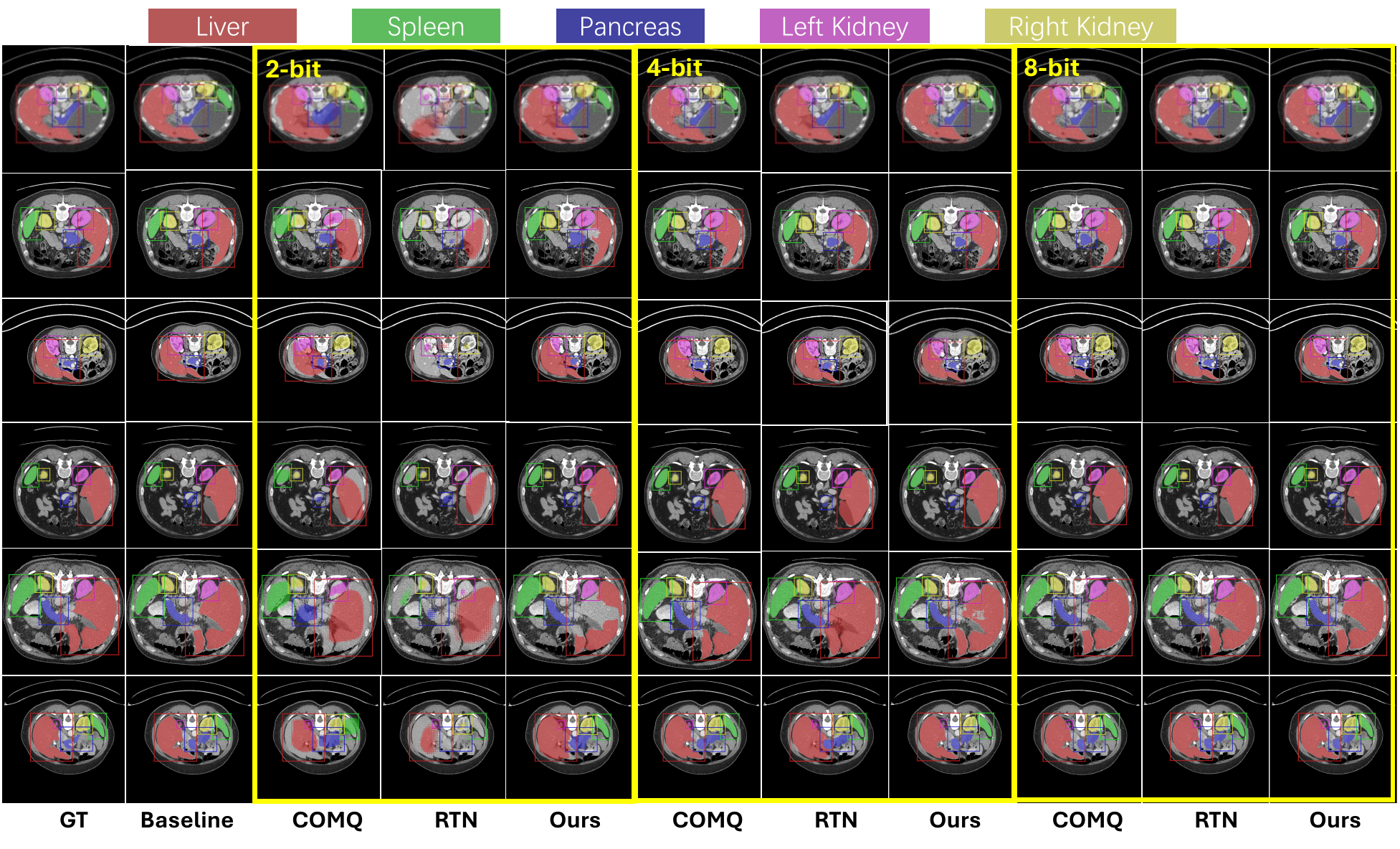}}

\caption{Conceptual illustration of COMQ and Permutation-COMQ.
\label{fig:compare}
\textbf{Top:} COMQ has quantization scale factors dominated by outliers due to heterogeneous weight magnitudes within each quantization unit, leading to coarse quantization of small weights. 
\textbf{Bottom:} Permutation-COMQ reorders weights by magnitude to group similar values, reducing intra-unit magnitude heterogeneity resulting in finer quantization resolution. The quantized weights are then mapped back to the original order via inverse permutation. \\}
\label{fig:summary}
\end{figure*}

\noindent{\textbf{Comparing with the Existing Methods}}
We evaluate the proposed permutation-COMQ method under different weight bit-width settings (e.g., 8-, 4-, and 2-bit) and compare it against baseline methods including RTN and COMQ. 

The experimental results are shown in Table~\ref{tab:ct}. The results demonstrate that as the weight-bit decreases, both the DSC and NSD scores for Permutation-COMQ significantly outperform existing approaches, namely COMQ and RTN. At 8 bits, Permutation-COMQ achieves a DSC of 93.615\% and an NSD of 93.204\%, surpassing COMQ (93.486\% and 92.938\%, respectively) and RTN (93.499\% and 92.936\%). These findings suggest that Permutation-COMQ maintains superior performance even with a reduced bit, indicating that its quantization strategy effectively preserves critical features despite lower computational resources. 
Our method consistently demonstrates better performance when comparing the results at 4-bit and 2-bit levels. For instance, at 4-bit quantization, Permutation-COMQ achieves a DSC of 93.434\%, outperforming COMQ (91.874\%) and RTN (90.526\%), while its NSD score of 93.089\% surpasses both COMQ (90.011\%) and RTN (86.755\%) in most cases. This trend persists at 2 bits, where Permutation-COMQ reaches a DSC of 86.939\% and a NSD of 78.935\%, notably exceeding the COMQ and RTN performance. 
In order to compare different quantization methods more intuitively, we visualize the predicted masks. From Fig. \ref{fig:compare}, we can see that it is consistent with our previous analysis.
CVPR

\subsubsection{Ablation Study} 
To demonstrate the impact of the proposed Per-Channel Weight-Aware approach on the entire quantization process, we conducted ablation experiments at 2-bit, 4-bit, and 8-bit quantization levels. 

The results are presented in Table \ref{tab:ab}. It is evident that directly applying quantization without considering the varying distributions of weights across different sizes significantly reduces segmentation accuracy. This effect is particularly pronounced at the 4-bit level, where NSD only reaches 49.404\%. Meanwhile, when incorporating Per-Channel Weight-Aware, as shown in Table 1, our method not only preserves accuracy but even improves it. This further validates the superiority of our approach.

\section{Conclusion}
In this paper, we introduce a powerful quantization method, Permutation-COMQ, to address the deployment challenges encountered by current medical image analysis foundation models on resource-limited medical devices. In contrast to recent approaches that rely on back-propagation or the estimation of the Hessian inverse to minimize reconstruction error, Permutation-COMQ addresses a series of univariate quadratic minimization problems, each of which has a closed-form solution. This approach eliminates the need for back-propagation, relying instead on dot products and rounding operations, and does not require any hyperparameters. Furthermore, we rearrange the weight matrix using a weight-aware strategy. This approach addresses the accuracy loss induced by channel-wise scaling during quantization, minimizing the sacrifice in model performance. 

In future work, we plan to further explore the combination of quantization with other optimization approaches, aiming to establish a paradigm for improving the performance of large medical models.

{
    \small
    \bibliographystyle{ieeenat_fullname}
    \FloatBarrier
    \bibliography{main}
}

\end{document}